\documentclass{Interspeech2024}
\usepackage[caption=false,font=normalsize,labelfont=sf,textfont=sf]{subfig}
\usepackage{mathtools}
\usepackage{verbatim}
\usepackage{multirow}
\usepackage{tabularx}
\usepackage{caption}
\usepackage[table]{xcolor}
\usepackage{makecell}
\definecolor{lightgray}{rgb}{0.8, 0.8, 0.8}

\interspeechcameraready

\title{An Initial Investigation of Language Adaptation for TTS Systems under Low-resource Scenarios \vspace{-0.8cm}}

\name[affiliation={1,2}]{Cheng}{Gong}
\name[affiliation={5}]{Erica}{Cooper}
\name[affiliation={2}]{Xin}{Wang}
\name[affiliation={1}]{Chunyu}{Qiang}
\name[affiliation={3}]{Mengzhe}{Geng}
\name[affiliation={4}]{Dan}{Wells}
\name[affiliation={1,*}]{Longbiao}{Wang}
\name[affiliation={1}]{Jianwu}{Dang}
\name[affiliation={3}]{Marc}{Tessier}
\name[affiliation={3}]{Aidan}{Pine}
\name[affiliation={4}]{Korin}{Richmond}
\name[affiliation={2}]{Junichi}{Yamagishi}

\address{\vspace{-0.4cm}
  $^1$Tianjin Key Laboratory of Cognitive Computing and Application,\\
College of Intelligence and Computing, Tianjin University, Tianjin, China\\
  $^2$National Institute of Informatics, Japan $^3$National Research Council Canada \\
  $^4$The Centre for Speech Technology Research, University of Edinburgh \\
  $^5$National Institute of Information and Communications Technology, Japan
  \vspace{-0.1cm}}
\email{\{gongchengcheng, longbiao\_wang\}@tju.edu.cn}

\keywords{multilingual speech synthesis, low-resource, language adaptation, fine-tuning}

\begin{document}
\maketitle
\author{\thanks{The maximum number of authors}}
\begin{abstract}
Self-supervised learning (SSL) representations from 
massively multilingual models offer a promising solution for low-resource language speech tasks. Despite advancements, language adaptation in TTS systems remains an open problem. This paper explores the language adaptation capability of ZMM-TTS, a recent SSL-based multilingual TTS system proposed in our previous work.
We conducted experiments on 12 languages using limited data with various fine-tuning configurations.
We demonstrate that the similarity in phonetics between the pretraining and target languages, as well as the language category, affects the target language's adaptation performance.
Additionally, we find that the fine-tuning dataset size and number of speakers influence adaptability. Surprisingly, we also observed that using paired data for fine-tuning is not always optimal compared to audio-only data. Beyond speech intelligibility, our analysis covers speaker similarity, language identification, and predicted MOS.

\end{abstract}

\section{Introduction}
Neural text-to-speech (TTS) \cite{ren2021fastspeech,kim2021conditional,tan2024naturalspeech} models have made remarkable progress in many industrial applications and academic research. However, most previous multilingual and multispeaker TTS models \cite{DBLP:conf/interspeech/ZhangWZWCSJRR19,badlani23_interspeech,DBLP:conf/icml/CasanovaWSJGP22} are still limited in supporting a wide range of languages and speakers, as they require a large amount of high-quality training data. Preparing data resources and building TTS systems for a specific language can be costly,
not to mention replicating this process for thousands of languages globally.
\renewcommand{\thefootnote}{}
\footnotetext{* denotes the corresponding author.}
\renewcommand{\thefootnote}{1}

With the advent of massively multilingual speech models like XLSR \cite{DBLP:journals/corr/abs-2006-13979}, using self-supervised learning (SSL) speech representations from multilingual models has become a promising solution for low-resource language speech processing tasks.
Considering TTS tasks as downstream tasks for SSL models, \cite{siuzdak2022wavthruvec} introduced WavThruVec, a two-stage architecture that employed high-dimensional wav2vec 2.0 embeddings as an intermediate speech representation.
A multilingual and multispeaker two-stage framework called ZMM-TTS was proposed in our prior work for low-resource language speech synthesis, which uses quantized latent speech representations from self-supervised model XLSR \cite{DBLP:journals/corr/abs-2006-13979}.
We also tested the efficiency on two hypothetical low-resource languages (Italian and Polish).
In \cite{jeong2024transfer}, the authors leveraged self-supervised speech representations as intermediate linguistic representations and
found increased efficiency in low-resource, multi-lingual, and zero-shot multi-speaker TTS tasks with a small amount of paired text and speech data supervised learning. 

In addition to speech-based SSL models, recent works confirm that using text pretraining also helps improve low-resource language TTS.
For example, in ZMM-TTS we used a large-scale pre-trained multilingual language model XPhoneBERT \cite{xphonebert} for phoneme representations and directly employed this model as an input phoneme encoder of TTS. Our experimental results show that XPhoneBERT helps boost the performance of low-resource languages. In \cite{saeki2023learning}, the authors first perform masked language model pretraining with multilingual text-only data. They then train the model with paired data in a supervised manner. The evaluation results of this study demonstrate highly intelligible zero-shot TTS for an unseen language.

Leveraging the self-supervised models trained on massive datasets, speech synthesis systems can generate the voices of specific languages and speakers with limited training data. However, most of these models have only been tested on Indo-European languages such as Italian, Spanish, and German, with only one or two low-resource languages being tested. Significant pronunciation differences may exist among various languages, such as Chinese and English. Mandarin Chinese is a tonal language, while English is not. Considering there are over 7,000 languages worldwide,
it is worthwhile to investigate the effectiveness of utilizing self-supervised models for achieving low-resource speech synthesis across various languages. 
Furthermore, the current method often straightforwardly involves fine-tuning with limited data to achieve low-resource language speech synthesis. Whether paired data or audio-only data is the most effective fine-tuning approach remains unclear. It is also unclear how the size of fine-tuning datasets and the number of speakers included affect the final performance of speech synthesis. 
Moreover, most previous works have only explored the intelligibility of synthesized speech in low-resource languages, neglecting a comprehensive comparison of other metrics such as speaker similarity and language identification.

This study examines the questions listed above using the ZMM-TTS system and presents the following contributions:
\begin{itemize}
\item [1.] We analyzed the adaptation of 12 different languages for the SSL-based TTS system in low-resource scenarios and found that adaptation performance is correlated with phonetic similarity and language category (tonal or non-tonal language). 
\item [2.] We conducted several fine-tuning experiments to analyze how the dataset size, number of speakers, and text inclusion in training data affected final adaptation performance.
\item [3.] We evaluated language adaptation performance using multiple metrics, including speech intelligibility, speaker similarity, language identification performance, and predicted MOS. Additionally, we analyzed the potential relationships between different metrics.
\end{itemize}
To our knowledge, this is the first study conducted on a large number of languages and fine-tuning configurations on SSL based TTS model. 
We strongly recommend that readers listen to the samples on the demo page at \url{https://nii-yamagishilab.github.io/samples-ZMM-TTS-series/}. The results are reproducible, and code will be released on \url{https://github.com/nii-yamagishilab/ZMM-TTS}. 

\section{Multilingual TTS system}
Our end-to-end multilingual and multi-speaker text-to-speech model architecture is based on ZMM-TTS. This section outlines the architecture and describes the text representations we used for this study.

\noindent\textbf{Model architecture.} ZMM-TTS consists of a text-to-discrete
speech representations module (\texttt{txt2vec}) and a representations-to-waveform module (\texttt{vec2wav}).
The \texttt{txt2vec} module is based on FastSpeech2 \cite{ren2021fastspeech} which includes token embeddings, an encoder, and a decoder to predict discrete SSL representations. We use a learnable aligner \cite{badlani2021OneTTSAlignment} to learn phoneme duration.
For the \texttt{vec2wav} module, ZMM-TTS adopts a 2-stage multi-head encoder as \cite{guo2022towards,guo2022multi} to convert discrete SSL representations to waveforms.

\textbf{Pretrained phoneme representations.}
In our previous work, we experimented with three different text representations: characters, phonemes (IPA) and pre-trained phoneme representations. 
In this study, we focus only on pre-trained phoneme representations, as ZMM-TTS has already demonstrated their advantages in low-resource scenarios. Additionally, the pre-trained phoneme representations based on XPhoneBERT \cite{xphonebert} support all the languages in our experimental setup.
XPhoneBERT is built on BERT-base architecture, pre-trained on RoBERTa\footnote{https://huggingface.co/docs/transformers/model\_doc/roberta} with 330M sentences and phoneme transcriptions across 100+ languages. 
Here, XPhoneBERT is used as the input phoneme encoder instead of the transformer architecture used in FastSpeech2.\renewcommand{\thefootnote}{2}
To prepare the data for XPhoneBERT, we first need to convert text sentences into a sequence of phonemes. We use the CharsiuG2P toolkit\footnote{https://github.com/lingjzhu/CharsiuG2P} for this purpose, as XPhoneBERT relies on phoneme-level input labels. \renewcommand{\thefootnote}{3}
During our finetuning experiment, the sequence of phonemes is converted into hidden representations using the pre-trained XPhoneBERT base model\footnote{https://huggingface.co/vinai/xphonebert-base}, whose parameters are also updated.

\vspace{-0.2cm}
\section {Language adaptation experiment}
\subsection{Datasets}
\textbf{Pretraining languages}: The pretraining data is the same as \cite{gong2023zmm}, which includes data from six languages: 
\textit{English (eng), French (fra), German (deu), Portuguese (por), Spanish (spa), Swedish (swe).}

\textbf{Adaptation languages}: 
We investigated 12 languages to analyze low resource adaptation performance: 
\textit{Bulgarian (bul), Croatian (hrv), Czech (ces), Dutch (nld), Italian (ita), Japanese (jpn), Korean (kor), Chinese (cmn), Polish (pol), Russian (rus), Turkish (tur), Vietnamese (vie).}
The Italian and Dutch are from the MLS dataset \cite{Pratap2020MLSAL}, the Chinese dataset is obtained from AISHELL-3 \cite{AISHELL-3_2020}, and data for other languages is derived from the GlobalPhone \cite{schultz2013globalphone} dataset.
It should be noted that all 12 languages are covered in the XPhoneBERT \cite{xphonebert} model, but there still are five languages \textit{(Bulgarian, Croatian, Czech, Japanese, Korean)} that are not covered in the XLSR \cite{DBLP:journals/corr/abs-2006-13979} model, and their language adaptation performance may be affected by their absence in XLSR.
We resampled all audio to 16 kHz sampling rate and applied amplitude normalization
using sv56 \cite{sv56}.

\begin{table}
    \centering
    \setlength\tabcolsep{2.5pt}
    \caption{Fine-tuning data set configurations. S, M, and L denote small, medium, large.}
    \label{tab:dataszie}
    \begin{tabular}{l|cccc|cccc|cccc}
        \toprule
        Name & S1 &S2 &S3  &S4  &M1 &M2 &M3 &M4 &L1 &L2 &L3 &L4    \\ \hline
        Spk  & 2 &4 &10 &20 &2 &4 &10 &20 &2 &4 &10 &20 \\ \hline
        Utt  &12 &6 &2 &1 &25 &12 &5 &2 &50 &25 &10 &5 \\ \hline
        Total &24 &24 &20 &20 &50 &48 &48 &40 &100 &100 &100 &100 \\
        \bottomrule
    \end{tabular}
    \vspace{-0.4cm}
\end{table}

\subsection{Fine-tune method}
\renewcommand{\thefootnote}{4}
We firstly pretrained the ZMM-TTS\footnote{https://github.com/nii-yamagishilab/ZMM-TTS} model on six languages, then finetune this model on 12 different languages.
To assess language adaptation in low-resource settings, we use a fine-tuning dataset of no more than 100 paired data samples for each language.
In order to analyze the impact of the number of speakers and the total amount of utterances from the fine-tuning data set on the final adaptation performance, we employed various configurations of fine-tuning data size as shown in Table \ref{tab:dataszie}.
Given the adaptation data sets listed in Table 1,  we also compared two different fine-tuning methods and included a zero-shot testing scenario: 
\begin{itemize}
\item
\textbf{Paired data fine-tuning.} We used paired data $\{\text{text}, \text{audio}\}$ and performed fine-tuning on both the \texttt{txt2vec} and \texttt{vec2wav} models.
\item 
\textbf{Audio-only fine-tuning.} We used audio-only data for fine-tuning the \texttt{vec2wav} model, and during testing, \texttt{txt2vec} processes the input in a zero-shot manner.
\item
\textbf{Zero-shot.} Without employing any data for fine-tuning, both \texttt{txt2vec} and \texttt{vec2wav} were directly tested on zero-shot inference.
\end{itemize}

For each language, we were able to conduct a comparison on a total of 25 configurations, including 12 (data sizes) $\times$ 2 (paired and unpaired) fine-tuned methods with limited data and one zero-shot model. For the same size of fine-tuning data, we use superscript to distinguish between different fine-tuning approaches. $\{S1,S2,\cdots,L4\}$ represents audio-only fine-tuning, while $\{S1^{\prime},S2^{\prime},\cdots,L4^{\prime}\}$ represents paired data fine-tuning. In the subsequent sections, we use \texttt{0} to represent zero-shot inference.
All fine-tuned models were trained on an A100 GPU with an effective batch size of 16 and trained for 600 epochs. 
\subsection{Evaluation metrics}
Due to the time and costs of subjective tests, it is difficult to recruit enough speakers to evaluate each language. As such, we used the following objective evaluation:
\begin{itemize}
\setlength{\itemindent}{0em}
\item
\textbf{Character error rate (CER).} 
\renewcommand{\thefootnote}{5}
We objectively evaluate performance by measuring the intelligibility of speech content using Whisper\footnote{https://github.com/openai/whisper}. 
We synthesized 100 sentences for each language and computed the CER between the input text and the ASR-produced transcripts.
\item 
\textbf{Language identification probability (LI).}
Similar to CER,  Whisper will also recognize the probability that these utterances belong to the target language. It is important to note that Whisper's predictions regarding text and language are two independent processes. During text prediction process, Whisper uses the ground-truth language ID as input. 
\item
\textbf{Speaker Encoder Cosine Similarity (SECS).} 
\renewcommand{\thefootnote}{6}
To assess the similarity between the synthesized voice and the original speaker, we determine the SECS by measuring the cosine similarity between the speaker embeddings of two audio samples extracted from the speaker encoder through Resemblyzer package\footnote{https://github.com/resemble-ai/Resemblyzer}.
For each language, we use the same 2 (1 female, 1 male) seen speakers and 4 (2 female, 2 male) unseen speakers for the speaker similarity test, and three sentences for each speaker.
\item
\textbf{UT-MOS}.
\renewcommand{\thefootnote}{7}
We employed automatic MOS (UT-MOS) prediction model\footnote{https://github.com/sarulab-speech/UTMOS22} to assess naturalness as \cite{saeki2023learning,saeki2022utmos}.
\end{itemize}
UT-MOS, CER, and LI were measured on the same test set.
\subsection{Language similarity analysis}

Inspired by the use of angular similarity (calculable from cosine similarity) between two languages’ vectors of phone frequencies to measure the similarity between their phone systems \cite{do2022text,do2023strategies}, we followed this method to analyze the phonetic similarity between 12 adaptation languages and six pretraining languages in our study.
For language A, we extracted its phone set through CharsiuG2P and then computed its vector of phone frequencies $\text{PF}_A$. Then, for the similarity between languages A and B, $S_{A,B}$ between $\text{PF}_A$ and $\text{PF}_B$ was calculated as follows:
\begin{equation}
S_{A,B} = 1-\frac{2}{\pi} \arccos(\frac{\text{PF}_A^\top \text{PF}_B}{\lVert \text{PF}_A \rVert \lVert \text{PF}_B \rVert}) 
\end{equation}

This metric, known as Angular Similarity of Phone Frequencies (ASPF), indicates the degree of similarity between two languages. The value of ASPF $S_{A,B}$ ranges from 0 to 1.
\section{Result and discussions}
\begin{figure}[t]
  \centering
  \includegraphics[width=\linewidth]{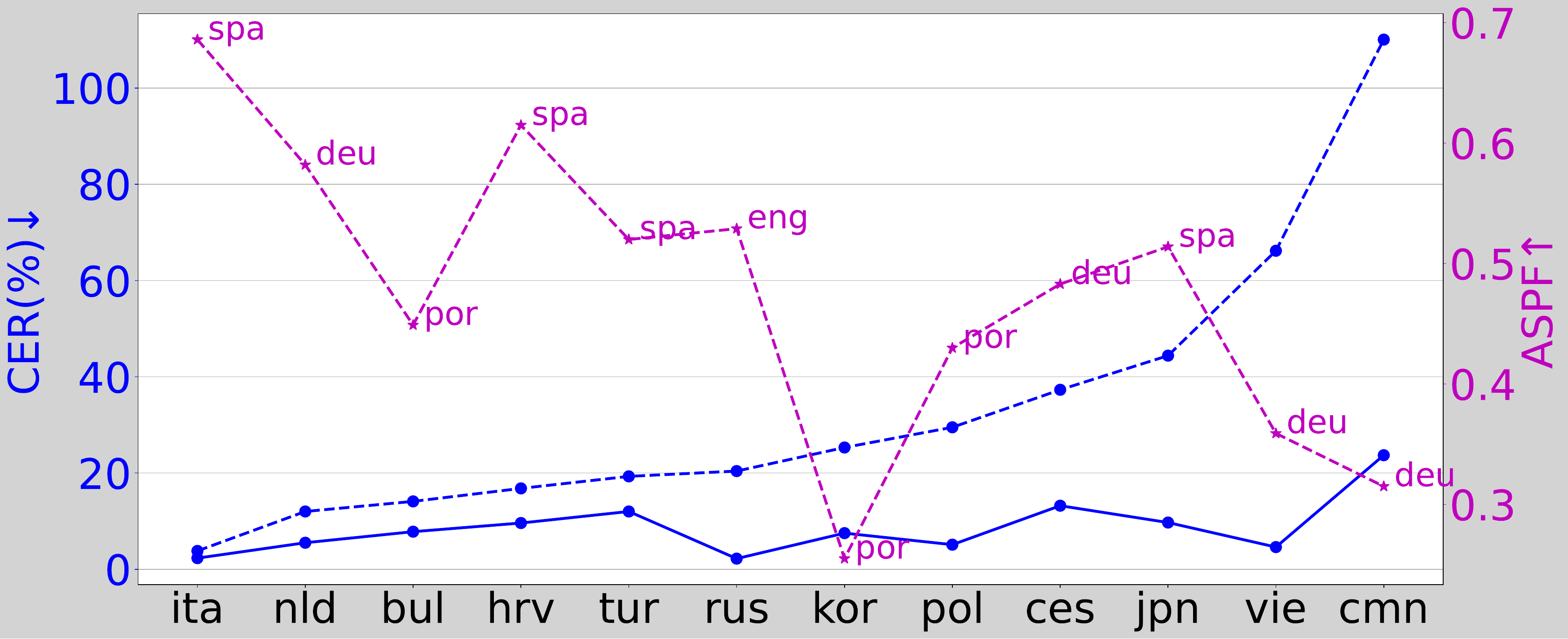}
  \caption{CER and ASPF values for different languages. The blue dashed line represents the best CER performance achievable by synthesized audio from 25 configurations, while the solid line represents the CER performance of natural audio. The purple dashed line represents the ASPF value most similar to the 6 pre-trained languages and its corresponding language.}
  \label{fig:1}
   \vspace{-0.6cm}
\end{figure}

\begin{figure}[t]
  \centering
  \includegraphics[width=0.9\linewidth]{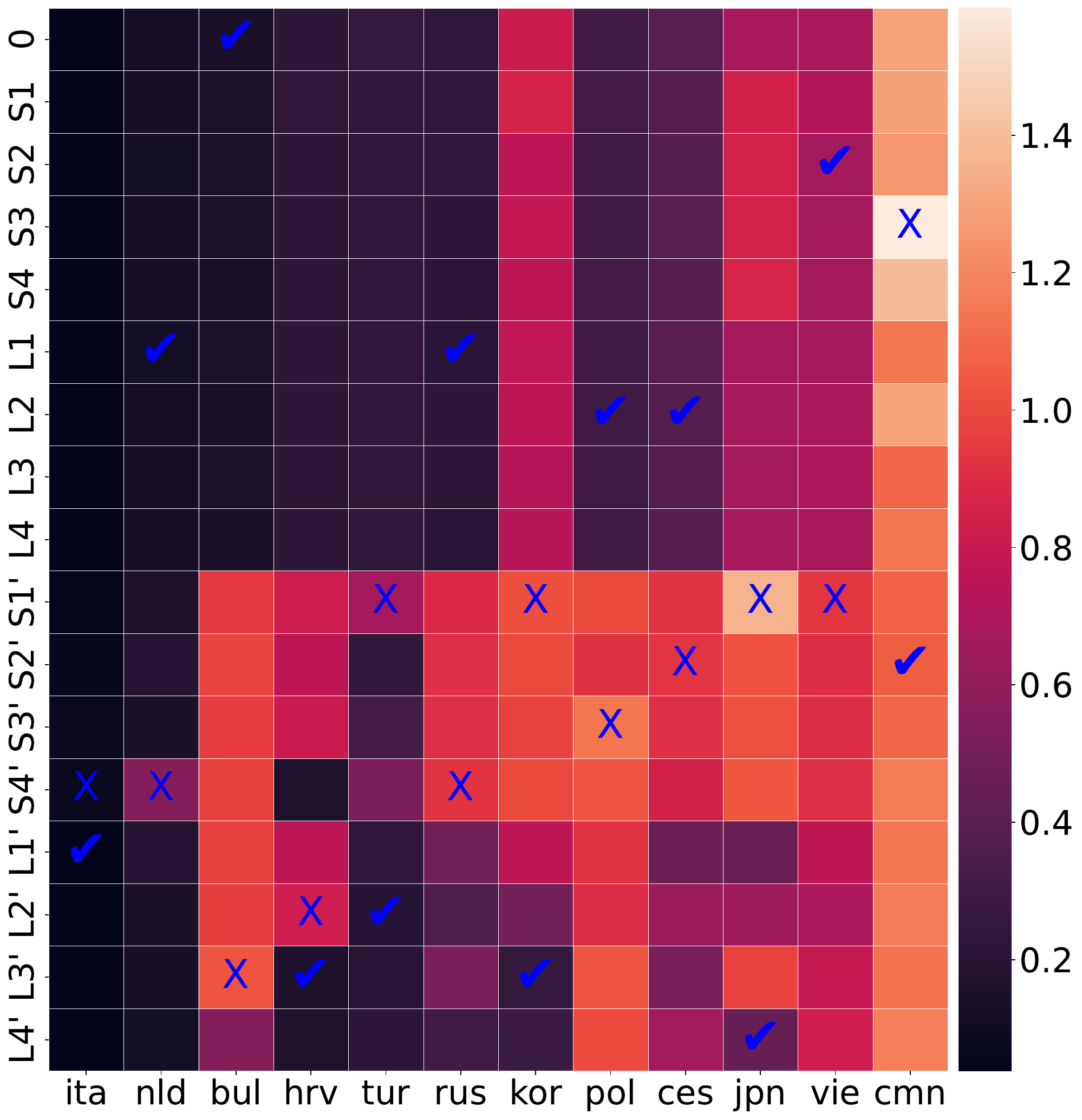}
  \caption{CER results for different languages under various fine-tuning methods. $\checkmark$ represents the best result for each language, while an $\times$ indicates the worst result.}
  \label{fig:2}
  \vspace{-0.6cm}
\end{figure}
\begin{table*}[t]
\centering
\setlength\tabcolsep{6pt}
\caption{Results of different metrics. Bold indicates the best result for each language under different fine-tune configurations.}
\label{tab:2}
\begin{tabular}{c|c|r|rrrr|rrrr|r}
\toprule
\multirow{3}{*}{Metrics}& \multirow{3}{*}{Lang} & \multicolumn{9}{c|}{Fine-tuning configurations} & \multirow{3}{*}{GT} \\ 
\cline{3-11}
& & Zero-shot& \multicolumn{4}{c|}{Audio-only} & \multicolumn{4}{c|}{Paired data} &  \\ 
\cline{3-11}
& & \scriptsize \texttt{$0$} & \scriptsize \texttt{$L1$}& \scriptsize \texttt{$L2$}& \scriptsize \texttt{$L3$} & \scriptsize \texttt{$L4$}& \scriptsize \texttt{$L1^{\prime}$}& \scriptsize \texttt{$L2^{\prime}$} & \scriptsize \texttt{$L3^{\prime}$} & \scriptsize \texttt{$L4^{\prime}$} & \\ \hline

\multirow{3}{*}{CER (\%) $\downarrow$} &ita & 4.6 &4.6 &4.7 &4.6 &4.8 &\textbf{3.8} &\textbf{3.8} &4.7 &4.3 &2.3 \\
& jpn & 67.9 & 67.0 &67.4 & 67.2 &67.8 & 45.4 & 64.5 &97.6 &\textbf{44.4} &9.7 \\
& tur &24.4 &23.5 & 22.7 & 23.1 &23.1 & 23.4 & \textbf{19.3} & 20.7 & 21.5 & 12.0 \\
\hline
\multirow{3}{*}{LI (\%) $\uparrow$} &ita &90.6 & 91.6 & 92.2 & 92.0 &92.9 &96.7&96.6&95.8&\textbf{97.0}&98.1 \\
& jpn & 41.2 & 59.1 &60.5 & 61.4 &62.0 &85.0 &92.2 &19.7 & \textbf{88.6} &97.7 \\
&tur & 26.1 &21.6 &25.9 &25.7 &26.2 &66.9 &\textbf{77.0} &74.9 &75.3 &98.4\\
\hline
\multirow{3}{*}{Seen SECS $\uparrow$} & ita &0.898 &0.924 &0.926 &0.913 &0.901 &0.948 &\textbf{0.951} & 0.930 & 0.904 & 0.999 \\
&jpn & 0.849 & 0.921 & 0.896 & 0.889 & 0.881 & \textbf{0.942} &0.917  &0.799 &0.906 &0.999\\ 
&tur &0.909 &\textbf{0.949} &0.942 &0.936 & 0.929 &\textbf{0.949} &0.937 &0.933 &0.916 &0.999 \\
\hline
\multirow{3}{*}{Unseen SECS $\uparrow$} &ita &\textbf{0.842} & 0.828 &0.841 &0.840 &0.832 & 0.803 &0.819 & 0.831 & 0.832 & 0.999 \\
&jpn &0.853 & 0.829 & 0.858 & \textbf{0.867} & 0.846 & 0.807 &0.851 &0.773 &0.847 &0.999 \\
&tur &0.876 &0.840 &0.864&0.862 &0.859 &0.845 &\textbf{0.877} & 0.872&0.873 &0.999 \\
\hline
\multirow{3}{*}{UT-MOS $\uparrow$} &ita & \textbf{3.359} & 3.227 &3.235 & 3.194 &3.227 & 3.120 & 3.145 & 3.065 & 3.121 &3.111 \\
&jpn & \textbf{3.286} &3.136 & 3.100 & 3.150 & 3.178 & 2.977 & 3.078 & 1.840 & 3.050 & 3.109 \\
&tur &\textbf{2.975} &2.812 &2.882 &2.899 &2.896 &2.723 &2.824 &2.851 &2.896 &2.971 \\
\bottomrule
\end{tabular}
\vspace{-0.6cm}
\end{table*}

\subsection{Impact of language variation}
Following the experimental setup, we performed fine-tuning and testing on 12 languages using models pretrained on six languages. 
Figure 1 reports the best intelligibility results (CER) among the 25 configurations achieved for each language.
Firstly, we have observed that there is a significant variation in the performance across different languages.
Fine-tuning with as little as 100 target-language utterances, most European languages, such as Italian, Dutch, and Bulgarian, can achieve good speech intelligibility (CER lower than 20\%).
However, for East Asian languages such as Chinese, Japanese, and Vietnamese the generated speech intelligibility is relatively lower (CER higher than 40\%).
Especially for Chinese, we consulted native speakers who also found it impossible to understand the content.
This result is as expected, as these East Asian languages and the six pre-trained languages have significant differences in pronunciation.

To further clarify the impact of the similarity between the target language and the languages included in pre-training on the final adaptation results, we calculated the ASPF and measured the similarity between the target language and each language included in the data set for pre-training. Higher ASPF values indicate more similarity to the language used for pre-training.
Figure \ref{fig:1} also shows the most similar language along with their ASPF values. 
Firstly, we found strong correlations ({$r=-0.630, p=0.028$}), using Pearson correlation coefficient (PCC), between the ASPF and the CER values. 
Italian has a high similarity (0.686) with Spanish in the pre-training corpus. As a result, its best CER results reflect this similarity. 
East Asian languages, such as Chinese, Vietnamese, and Korean, have lower similarity to the languages in the pre-training corpus.
This indicates that phonetic differences between the target language and the languages in pre-training may impact the final adaptation performance. 

While Japanese exhibits a high similarity with the Spanish languages in pre-training, it also has a higher CER, although we consulted native speakers who found synthesized Japanese audio to be generally understandable. 
Korean exhibits good intelligibility despite low similarity to languages in pre-training, which is opposite to Japanese. 
Further investigation is still needed on these two East Asian languages. 
In addition to phonetic similarity, differences in pronunciation, such as tonality, may affect language adaptation performance. 

We also observed a strong correlation (PCC $r=0.715, p=0.008$) between the CER of synthesized audio and natural audio. For instance, across 12 languages' natural speech, Chinese also achieved the poorest performance as synthesized speech.
\subsection{Impact of finetuning configurations}
Figure \ref{fig:2} illustrates the CER results for 12 different languages under 12 finetuneing configurations. The complete results for all 25 configurations can be found on the demo website.
From Figure \ref{fig:2}, we can make the following observations:
\begin{itemize}
\setlength{\leftmargin}{0pt}
\item  
With limited data, fine-tuning both of \texttt{txt2vec} and \texttt{vec2wav} is not always the best option. Using an extremely small amount of data, such as only 20 paired samples, can make the model prone to overfitting, compromising the adaptability of the pre-trained model. In Figure \ref{fig:2}, 9 languages (ita, nld, tur, rus, kor, pol, ces, jpn, vie) obtained the poorest CER results when fine-tuned with only about 20 paired samples.
\item  The difference in results between fine-tuning using only audio data and the zero-shot scenario is not much, which was expected because speech intelligibility in synthesized speech is influenced more by training text than audio.
\item Increasing the amount of data in the training set enhances the performance of fine-tuning. For example, the best results were obtained for 9 languages (ita, nld, hrv, tur, rus, kor, pol, ces, jpn) when fine-tuned with approximately 100 samples (paired or audio-only) in Figure \ref{fig:2}. 
\end{itemize}
\vspace{-0.3cm}
\subsection{Results across different metrics}
\vspace{-0.2cm}
To better analyze various metrics in low-resource language scenarios for speech synthesis, we list all metrics' results for each language under 9 configurations. Table \ref{tab:2} shows the results of three typical languages, including a European language (Italian), an East Asian language (Japanese), and a Central Asian language (Turkish).
As space is limited, the results for all metrics across all 25 configurations for the 12 languages are available on the demo webpage.
Next, we will primarily analyze the results from three perspectives: speaker similarity, language identification, and predicted MOS values.

\textbf{Speaker similarity.} 
For speaker similarity, fine-tuning with data from the target speaker improves speaker similarity of synthesized audio compared to unseen speakers.
Furthermore, with total training data unchanged, increasing the number of utterances from the target speaker in training data is more effective in enhancing the similarity than increasing the number of speakers.
In these three languages, both $L3^{\prime} (10 spk * 10 utt)$ and $L4^{\prime} (20 spk * 5 utt)$ are lower than  $L1^{\prime} (2 spk * 50 utt)$ and $L2^{\prime} (4 spk * 25 utt)$ in terms of seen speaker similarity.

From Table \ref{tab:2}, we also observed that the zero-shot and fine-tuned models have similar performance on unseen speaker similarity. 
That means fine-tuning with limited data containing new speakers does not significantly improve the model's generalization ability to generate unseen speakers. The generalization ability of the model to unseen speakers may be more dependent on the number of speakers in the pretraining data.

\textbf{Language identification.}
From Table \ref{tab:2}, we can find that for these three different languages, there is a correlation between the language identification performance and CER for synthesized speech. 
The synthesized speech sounds more like the target language, resulting in higher intelligibility.
For three languages in Table \ref{tab:2}, in most cases, the speech synthesized after fine-tuning with paired data sounds more like the target language compared to the speech synthesized after fine-tuning with audio-only data or through zero-shot methods.

\textbf{Predicted MOS.}
As a SOTA model,  UT-MOS achieved better prediction results in some languages with abundant resources, such as English \cite{cooper2023voicemos}. However, its performance in our experiments was not satisfactory. The UT-MOS model consistently predicts the highest MOS values for zero-shot generated audio, although their speech intelligibility may be significantly worse than the audio synthesized using a sufficient amount of paired data fine-tuning. This indicates that the current MOS automatic prediction models still need improvement in their ability to support different languages.
\section{Conclusions}
This paper explores the language adaptation ability of ZMM-TTS, an SSL-based multilingual speech synthesis system. Experiments on 12 languages with various fine-tuning configurations reveal the impact of phonetic similarity and language category on adaptation performance. Additionally, we find that the fine-tuning dataset size and speaker diversity influence adaptability. Surprisingly, using paired data for fine-tuning is not always optimal compared to audio-only data. Beyond speech intelligibility, our analysis covers speaker similarity, language identification, and predicted MOS. Additionally, we analyzed the potential relationships between different metrics.
\newpage
\textbf{Acknowledgments:} This work was conducted during the first author’s research stay at NII, Japan supported by the China Scholarship Council (CSC) No. 202206250146. 
This work was supported in part by the National Natural Science Foundation of China under Grant (U23B2053, 62176182), MEXT KAKENHI Grants (21H04906) and the National Research Council of Canada’s Ideation Fund: `Small teams – Big Ideas’.

\bibliographystyle{IEEEtran}
\bibliography{mybib}

\begin{thebibliography}{10}
\providecommand{\url}[1]{#1}
\csname url@samestyle\endcsname
\providecommand{\newblock}{\relax}
\providecommand{\bibinfo}[2]{#2}
\providecommand{\BIBentrySTDinterwordspacing}{\spaceskip=0pt\relax}
\providecommand{\BIBentryALTinterwordstretchfactor}{4}
\providecommand{\BIBentryALTinterwordspacing}{\spaceskip=\fontdimen2\font plus
\BIBentryALTinterwordstretchfactor\fontdimen3\font minus \fontdimen4\font\relax}
\providecommand{\BIBforeignlanguage}[2]{{%
\expandafter\ifx\csname l@#1\endcsname\relax
\typeout{** WARNING: IEEEtran.bst: No hyphenation pattern has been}%
\typeout{** loaded for the language `#1'. Using the pattern for}%
\typeout{** the default language instead.}%
\else
\language=\csname l@#1\endcsname
\fi
#2}}
\providecommand{\BIBdecl}{\relax}
\BIBdecl

\bibitem{ren2021fastspeech}
\BIBentryALTinterwordspacing
Y.~Ren, C.~Hu, X.~Tan, T.~Qin, S.~Zhao, Z.~Zhao, and T.-Y. Liu, ``{FastSpeech 2: Fast and High-Quality End-to-End Text to Speech},'' in \emph{International Conference on Learning Representations}, 2021. [Online]. Available: \url{https://openreview.net/forum?id=piLPYqxtWuA}
\BIBentrySTDinterwordspacing

\bibitem{kim2021conditional}
J.~Kim, J.~Kong, and J.~Son, ``{Conditional variational autoencoder with adversarial learning for end-to-end text-to-speech},'' in \emph{Proc. ICML}.\hskip 1em plus 0.5em minus 0.4em\relax PMLR, 2021, pp. 5530--5540.

\bibitem{tan2024naturalspeech}
X.~Tan, J.~Chen, H.~Liu, J.~Cong, C.~Zhang, Y.~Liu, X.~Wang, Y.~Leng, Y.~Yi, L.~He \emph{et~al.}, ``{Naturalspeech: End-to-end text-to-speech synthesis with human-level quality},'' \emph{IEEE Transactions on Pattern Analysis and Machine Intelligence}, pp. 1--12, 2024.

\bibitem{DBLP:conf/interspeech/ZhangWZWCSJRR19}
Y.~Zhang, R.~J. Weiss, H.~Zen, Y.~Wu, Z.~Chen, R.~J. Skerry{-}Ryan, Y.~Jia, A.~Rosenberg, and B.~Ramabhadran, ``{Learning to Speak Fluently in a Foreign Language: Multilingual Speech Synthesis and Cross-Language Voice Cloning},'' in \emph{Proc. Interspeech}.\hskip 1em plus 0.5em minus 0.4em\relax {ISCA}, 2019, pp. 2080--2084.

\bibitem{badlani23_interspeech}
R.~Badlani, R.~Valle, K.~J. Shih, J.~F. Santos, S.~Gururani, and B.~Catanzaro, ``{RAD-MMM}: Multilingual multiaccented multispeaker text to speech,'' in \emph{Proc. Interspeech}, 2023, pp. 626--630.

\bibitem{DBLP:conf/icml/CasanovaWSJGP22}
E.~Casanova, J.~Weber, C.~D. Shulby, A.~C. J{\'{u}}nior, E.~G{\"{o}}lge, and M.~A. Ponti, ``{{YourTTS}: Towards Zero-Shot Multi-Speaker {TTS} and Zero-Shot Voice Conversion for Everyone},'' in \emph{Proc. ICML}, vol. 162.\hskip 1em plus 0.5em minus 0.4em\relax {PMLR}, 2022, pp. 2709--2720.

\bibitem{DBLP:journals/corr/abs-2006-13979}
A.~Conneau, A.~Baevski, R.~Collobert, A.~Mohamed, and M.~Auli, ``Unsupervised cross-lingual representation learning for speech recognition,'' in \emph{Proc. {{Interspeech}}}, 2021, pp. 2426--2430.

\bibitem{siuzdak2022wavthruvec}
H.~Siuzdak, P.~Dura, P.~{van Rijn}, and N.~Jacoby, ``{{WavThruVec}}: {{Latent}} speech representation as intermediate features for neural speech synthesis,'' in \emph{Proc. {{Interspeech}}}, 2022, pp. 833--837.

\bibitem{jeong2024transfer}
M.~Jeong, M.~Kim, B.~J. Choi, J.~Yoon, W.~Jang, and N.~S. Kim, ``{Transfer Learning for Low-Resource, Multi-Lingual, and Zero-Shot Multi-Speaker Text-to-Speech},'' \emph{IEEE/ACM Transactions on Audio, Speech, and Language Processing}, pp. 1519--1530, 2024.

\bibitem{xphonebert}
L.~T. Nguyen, T.~Pham, and D.~Q. Nguyen, ``{XPhoneBERT: A Pre-trained Multilingual Model for Phoneme Representations for Text-to-Speech},'' in \emph{Proc. {{Interspeech}}}, 2023, pp. 5506--5510.

\bibitem{saeki2023learning}
T.~Saeki, S.~Maiti, X.~Li, S.~Watanabe, S.~Takamichi, and H.~Saruwatari, ``{Learning to Speak from Text: {Zero}-Shot Multilingual Text-to-Speech with Unsupervised Text Pretraining},'' in \emph{Proc. {{IJCAI}}}, 2023, pp. 5179--5187.

\bibitem{badlani2021OneTTSAlignment}
R.~Badlani, A.~{\L}a{\'n}cucki, K.~J. Shih, R.~Valle, W.~Ping, and B.~Catanzaro, ``One {TTS} alignment to rule them all,'' in \emph{2022 IEEE International Conference on Acoustics, Speech and Signal Processing}.\hskip 1em plus 0.5em minus 0.4em\relax IEEE, 2022, pp. 6092--6096.

\bibitem{guo2022towards}
H.~Guo, F.~Xie, X.~Wu, F.~K. Soong, and H.~Meng, ``{MSMC-TTS}: Multi-stage multi-codebook {VQ-VAE} based neural {TTS},'' \emph{IEEE/ACM Transactions on Audio, Speech, and Language Processing}, vol.~31, pp. 1811--1824, 2023.

\bibitem{guo2022multi}
H.~Guo, F.~Xie, F.~K. Soong, X.~Wu, and H.~Meng, ``A multi-stage multi-codebook {VQ-VAE} approach to high-performance neural {TTS},'' in \emph{Proc. Interspeech}.\hskip 1em plus 0.5em minus 0.4em\relax {ISCA}, 2022, pp. 1611--1615.

\bibitem{gong2023zmm}
C.~Gong, X.~Wang, E.~Cooper, D.~Wells, L.~Wang, J.~Dang, K.~Richmond, and J.~Yamagishi, ``{ZMM-TTS: Zero-shot Multilingual and Multispeaker Speech Synthesis Conditioned on Self-supervised Discrete Speech Representations},'' \emph{arXiv preprint arXiv:2312.14398}, 2023.

\bibitem{Pratap2020MLSAL}
V.~Pratap, Q.~Xu, A.~Sriram, G.~Synnaeve, and R.~Collobert, ``{MLS: A Large-Scale Multilingual Dataset for Speech Research},'' \emph{Proc. {{Interspeech}}}, pp. 2757--2761, 2020.

\bibitem{AISHELL-3_2020}
Y.~Shi, H.~Bu, X.~Xu, S.~Zhang, and M.~Li, ``{AISHELL-3: A Multi-speaker Mandarin TTS Corpus and the Baselines},'' in \emph{https://arxiv.org/abs/2010.11567}, 2015.

\bibitem{schultz2013globalphone}
T.~Schultz, N.~T. Vu, and T.~Schlippe, ``{Globalphone: A multilingual text \& speech database in 20 languages},'' in \emph{2013 IEEE International Conference on Acoustics, Speech and Signal Processing}.\hskip 1em plus 0.5em minus 0.4em\relax IEEE, 2013, pp. 8126--8130.

\bibitem{sv56}
\BIBentryALTinterwordspacing
I.~T. Union. {{Recommendation G.191}: Software Tools and Audio Coding Standardization}. (2005, Nov 11). [Online]. Available: \url{https://www.itu.int/rec/T-REC-P.56/en}
\BIBentrySTDinterwordspacing

\bibitem{saeki2022utmos}
T.~Saeki, D.~Xin, W.~Nakata, T.~Koriyama, S.~Takamichi, and H.~Saruwatari, ``{UTMOS: UTokyo-SaruLab system for VoiceMOS Challenge 2022},'' in \emph{Proc. Interspeech}, 2022, pp. 4521--4525.

\bibitem{do2022text}
P.~Do, M.~Coler, J.~Dijkstra, and E.~Klabbers, ``Text-to-speech for under-resourced languages: Phoneme mapping and source language selection in transfer learning,'' in \emph{Proceedings of the 1st Annual Meeting of the ELRA/ISCA Special Interest Group on Under-Resourced Languages}, 2022, pp. 16--22.

\bibitem{do2023strategies}
------, ``{Strategies in Transfer Learning for Low-Resource Speech Synthesis: Phone Mapping, Features Input, and Source Language Selection},'' in \emph{12th Speech Synthesis Workshop (SSW) 2023}, 2023, pp. 21--26.

\bibitem{cooper2023voicemos}
E.~Cooper, W.-C. Huang, Y.~Tsao, H.-M. Wang, T.~Toda, and J.~Yamagishi, ``{The VoiceMOS Challenge 2023: zero-shot subjective speech quality prediction for multiple domains},'' in \emph{2023 IEEE Automatic Speech Recognition and Understanding Workshop (ASRU)}.\hskip 1em plus 0.5em minus 0.4em\relax IEEE, 2023, pp. 1--7.

\end{thebibliography}

\end{document}